\documentclass[11pt]{report}
\usepackage{geometry, graphicx, kotex, imakeidx, titlesec, array} 
\usepackage{mathptmx}   
\usepackage{amsmath, amsthm, amssymb, mathrsfs, multirow, verbatim} 
\usepackage[nobiblatex]{xurl} 
\usepackage{indentfirst} 
\usepackage{booktabs} 
\usepackage{ragged2e}
\usepackage{caption}
\usepackage[table]{xcolor} 
\usepackage{setspace}
\usepackage{ragged2e}

\titleformat{\chapter}{\normalfont\huge\bfseries}{\chaptertitlename\ \thechapter.}{20pt}{\huge} 
\makeindex 

\begin{document}

\newpage   
\begin{center}
\pagenumbering{gobble} 
{\fontsize{16pt}{16pt}\selectfont Master's Thesis \par}  
\vspace{3cm} 
\huge Unified Low-Light Traffic Image Enhancement via Multi-Stage Illumination Recovery and Adaptive Noise Suppression
\par\vspace{3.5cm} 
{\fontsize{16pt}{16pt}\selectfont Siddiqua Namrah \par}  
\vspace{0.5cm}
{\fontsize{16pt}{16pt}\selectfont Department of Artificial Intelligence \par}  
\vspace{1.5cm}
{\fontsize{18pt}{18pt}\selectfont Graduate School \par} 
\vspace{0.5cm}
{\fontsize{18pt}{18pt}\selectfont Korea University \par} 
\vspace{1cm}
{\fontsize{14pt}{14pt}\selectfont February 2026} 
\end{center}

\newpage 
\begin{center}
\huge Unified Low-Light Traffic Image Enhancement via Multi-Stage Illumination Recovery and Adaptive Noise Suppression  
\par\vspace{2.0cm} 
{\fontsize{16pt}{16pt}\selectfont by \par Siddiqua Namrah \par}
\vspace{0.3cm}
\rule{.6\textwidth}{0.4pt} 
\par\vspace{0.2cm}
{\fontsize{16pt}{16pt}\selectfont under the supervision of Professor Seong-Whan
Lee \par}
\vspace{0.7cm}
{\fontsize{16pt}{18pt}\selectfont A thesis submitted in partial fulfillment of \par
 the requirements for the degree of \par Master of Science \par }   
\vspace{10pt}
{\fontsize{16pt}{16pt}\selectfont Department of Artificial Intelligence} 
\par\vspace{0.5cm}
{\fontsize{18pt}{18pt}\selectfont Graduate School \par \vspace{0.5cm}
 Korea University \par} 
\vspace{0.5cm}
{\fontsize{14pt}{14pt}\selectfont October 2025} 
\end{center}


\newpage 
\begin{center}
\vspace{1cm}
{\fontsize{16pt}{18pt}\selectfont
The thesis of Siddiqua Namrah has been approved \par
by the thesis committee in partial fulfillment\par
of the requirements for the degree of \par
Master of Science \par}  
\vspace{1cm}
{\fontsize{14pt}{14pt}\selectfont December 2025} 
\par\vspace{3cm}
\rule{.6\textwidth}{0.4pt}\par 
{\fontsize{16pt}{16pt}\selectfont Committee Chair: Seong-Whan Lee \par}
\vspace{1cm}
\rule{.6\textwidth}{0.4pt}\par 
{\fontsize{16pt}{16pt}\selectfont Committee Member: Wonzoo Chung \par}
\vspace{1cm}
\rule{.6\textwidth}{0.4pt}\par 
{\fontsize{16pt}{16pt}\selectfont Committee Member: Tae-Eui Kam \par}
\end{center}

\newpage 
\pagenumbering{roman} 
\newgeometry{paper=b5paper, left=20mm, right=20mm, top=30mm, bottom=30mm} 
\addcontentsline{toc}{chapter}{Abstract}
\begin{center}
\LARGE Unified Low-Light Traffic Image Enhancement via Multi-Stage Illumination Recovery and Adaptive Noise Suppression
\par\vspace{20pt}

\normalsize 
\doublespacing
by Siddiqua Namrah\par 
Department of Artificial Intelligence\par
under the supervision of Professor Seong-Whan Lee 
\par\vspace{20pt}
\large \textbf{Abstract}
\end{center}

\normalsize
\justifying 
\doublespacing

Enhancing low-light traffic imagery is a critical requirement for achieving reliable perception in autonomous driving, intelligent transportation, and urban surveillance systems. Traffic scenes captured under nighttime or dimly lit conditions often suffer from complex visual degradations arising from insufficient illumination, sensor noise amplification, motion-induced blur, non-uniform lighting, and strong glare generated by vehicle headlights or street lamps. These challenges significantly hinder downstream tasks such as object detection, tracking, and scene understanding, demonstrating the need for a robust enhancement framework capable of restoring visibility in diverse real-world conditions without relying on paired training data.

To address these issues, we propose a unified unsupervised deep learning framework specifically designed for low-light traffic image enhancement. The proposed model adopts a multi-stage architecture that decomposes the input into illumination and reflectance components and progressively refines each through three functionally specialized modules. First, an Illumination Adaptation module adjusts global and local brightness by leveraging contextual priors, ensuring natural exposure correction and contrast enhancement. Second, a Reflectance Restoration module equipped with joint spatial–channel attention suppresses noise while selectively recovering structural details crucial for traffic perception tasks. Third, an Over-Exposure Compensation module focuses on reconstructing saturated or washed-out regions caused by intense artificial light sources, effectively balancing luminance across the scene.

The overall framework operates under a fully unsupervised setting, guided by self-supervised reconstruction losses, reflectance smoothness priors, perceptual consistency constraints, and domain-aware regularization. These strategies enable the network to learn meaningful enhancement without ground-truth supervision while preserving textures, suppressing noise, and maintaining color fidelity. Extensive experiments on both general-purpose low-light datasets and traffic-oriented benchmark datasets demonstrate that the proposed method outperforms state-of-the-art techniques across PSNR, SSIM, LPIPS, and and no-reference metrics, including NIQE, and MetaIQA metrics. Qualitative evaluations further show that our approach produces clearer, more artifact-free reconstructions, improving the reliability of subsequent machine vision modules in real-world driving and surveillance environments.

Our findings indicate that the proposed unsupervised multi-stage framework offers a scalable and effective solution for enhancing low-light traffic scenes, contributing to safer and more reliable urban vision systems.\par

\par\vspace{20pt}


\newpage 
\begin{center}
\LARGE 야간 교통 영상 향상을 위한 다단계 조명 복원 및 잡음 억제 기반 딥러닝 프레임워크 
\par\vspace{20pt}
\normalsize 남 라  싯 디 카\par 
인 공 지 능 학 과\par 
지 도 교 수: 이 성 환
\par\vspace{20pt}
\addcontentsline{toc}{chapter}{국문초록}
\large \textbf{국문 초록}
\end{center}

\normalsize 

저조도 환경에서 획득된 교통 영상은 자율주행, 지능형 교통 시스템, 도시 감시와 같은 다양한 비전 기반 응용에서 핵심적인 역할을 수행함에도 불구하고, 조명 부족으로 인해 심각한 품질 저하가 발생한다. 야간 또는 어두운 환경에서 촬영된 교통 영상은 조도 불균형, 센서 기저 노이즈, 동적 객체에 의한 움직임 블러, 차량 전조등으로 인한 강한 눈부심 등 복합적이고 비선형적인 왜곡이 동시에 나타나기 때문에, 이를 효과적으로 복원하는 것은 기존의 단일 단계 또는 단순한 강화 기법으로는 해결하기 어렵다.

본 논문에서는 이러한 복잡한 저조도 교통 영상의 특성을 고려하여, 비지도 학습 기반의 통합 딥러닝 프레임워크를 제안한다. 제안된 모델은 입력 영상을 조명(illumination)과 반사(reflectance) 성분으로 분해한 뒤, 각각의 성분을 단계별로 보정·복원하기 위해 설계된 세 가지 다단계 모듈을 포함한다.

첫째, 조명 적응(IA) 모듈은 주변 문맥 정보를 활용해 밝기와 대비를 균형 있게 조절하며 전반적인 시각적 일관성을 확보한다. 둘째, 반사 복원(RR) 모듈은 공간 및 채널 주의 메커니즘을 통합하여 노이즈를 효과적으로 억제하고 객체의 경계 및 구조적 세부 정보를 정교하게 복원한다. 셋째, 과노출 보정(OEC) 모듈은 차량 전조등·가로등 등 강한 인공광에 의해 발생하는 국소적 포화 영역을 복구하여 실제 교통 환경에서 빈번히 나타나는 노출 불균형 문제를 완화한다.

또한 본 연구는 별도의 GT(ground truth)를 요구하지 않는 자가 지도(un-supervised) 학습 전략을 적용하여 다양한 조명 조건에 대한 일반화를 강화하였으며, 교통 영상의 특수성을 반영한 도메인 맞춤형 설계를 통해 노이즈 제거, 텍스처 보존, 조명 복원 사이의 균형을 효과적으로 달성하였다.

다양한 공용 저조도 데이터셋과 교통 환경 특화 데이터셋에서 수행한 광범위한 실험 결과, 제안 기법은 PSNR, SSIM, LPIPS와 같은 기준뿐만 아니라 NIQE, MetaIQA와 같은 무참조 지표에서도 최신 기법들을 능가하는 성능을 보였다. 특히 실제 야간 주행 및 도시 감시 환경에서의 강인한 복원 성능을 보여, 향후 교통 기반 지능형 시각 시스템의 핵심 모듈로 활용될 수 있음을 입증하였다.

\par\vspace{20pt}
\textbf{중심어}: 저조도 영상 향상, 교통 장면 복원, 비지도 학습, 다단계 처리, 조명-반사 분해, 주의 메커니즘, 노출 보정, 도시 시각 인식 시스템


\newpage
\chapter*{Preface}
\addcontentsline{toc}{chapter}{Preface}
\normalsize
This dissertation is submitted for the degree of Master of Science in Artificial Intelligence at Korea University. All of the research presented herein was conducted in the Department of Artificial Intelligence under the supervision of Professor Seong-Whan Lee from September 2023 to August 2025. The dissertation presents the core findings of my research on low-light traffic image enhancement. A portion of this work has been submitted to peer-reviewed journals, and the corresponding chapters retain conceptual and methodological similarities to the submitted manuscripts. However, the dissertation provides additional explanations, extended experimental results, and supplementary analyses to satisfy academic requirements and offer a comprehensive account of the study.

I was the lead investigator for all major aspects of this research, including conceptual development, model design, dataset construction, experimentation, and manuscript preparation. Neither this dissertation, nor any substantially similar work, has been submitted for any other degree or qualification at any institution. Part of this work has also been accepted for presentation at the 2025 IEEE International Conference on Systems, Man, and Cybernetics (SMC), Vienna, Austria.

\par\vspace{30pt}

\newpage
\chapter*{Acknowledgment}
\addcontentsline{toc}{chapter}{Acknowledgment}

This research was supported by the Institute of Information Communications
Technology Planning Evaluation (IITP) grant, funded by the Korea government
(MSIT) (No. RS-2019-II190079 (Artificial Intelligence Graduate School Program
(Korea University), No. IITP-2025-RS-2024-00436857 (Information Technology
Research Center (ITRC), and No. IITP-2025-RS-2025-02304828 (under the artificial intelligence star fellowship support program to nurture the best talents)).

\renewcommand*\contentsname{Table of Contents}
\addcontentsline{toc}{chapter}{Table of Contents}
\tableofcontents

\bigskip

\listoftables
\addcontentsline{toc}{chapter}{List of Tables}

\bigskip
A list of tables shall be included when there are tables in the thesis/dissertation. Table numbering can be continuous throughout the thesis/dissertation or by chapter (e.g., 1.1, 1.2, 2.1, 2.2...).

\listoffigures
\addcontentsline{toc}{chapter}{List of Figures}

\bigskip
List of figures should be prepared when figures are included in the thesis/dissertation. Figure numbering can be continuous throughout the thesis/dissertation or by chapter (e.g., 1.1, 1.2, 2.1, 2.2...).

\chapter*{Nomenclature}
\addcontentsline{toc}{chapter}{Nomenclature} 

\section*{Variables}
\begin{tabular}{ll}
$ I_\text{in} $ & Input low-light image \\
$ I_\text{out} $ & Enhanced output image \\
$ R $ & Reflectance component \\
$ L $ & Illumination component \\
$ \hat{L} $ & Estimated illumination \\
\end{tabular}

\section*{Metrics}
\begin{tabular}{ll}
PSNR & Peak Signal-to-Noise Ratio \\
SSIM & Structural Similarity Index \\
LPIPS & Learned Perceptual Image Patch Similarity \\
NIQE & Naturalness Image Quality Evaluator \\
MetaIQA & Meta-Learning based No-Reference Image Quality Assessment \\
\end{tabular}

\section*{Datasets}
\begin{tabular}{ll}
BDD100K-N & BDD100K Nighttime subset \\
LoLI-Street & Low-Light Street dataset \\
MVLT & Multi-View Low-Light Traffic dataset \\
\end{tabular}

\section*{Subscripts / Notation}
\begin{tabular}{ll}
$i$ & Pixel index \\
$c$ & Color channel \\
$s$ & Scale or spatial level \\
\end{tabular}

\section*{Abbreviations}
\begin{tabular}{ll}
LLIE & Low-Light Image Enhancement \\
CG & Channel-Guidance Module \\
CE & Color Enhancement Module \\
OEC & Over-Exposure Correction Module \\
GAN & Generative Adversarial Network \\
\end{tabular}

\bigskip
If nomenclature or a list of symbols is used, a section describing subscripts and abbreviations can be included.

\chapter{Introduction}\label{chap:intro}
\pagenumbering{arabic} 
The ability to perceive road environments reliably during nighttime is a foundational requirement for autonomous vehicles, advanced driver assistance systems, and modern traffic monitoring infrastructures. As these systems increasingly depend on vision-based interpretation for localization, scene understanding, and safety-critical decision making, the limitations imposed by low-light conditions pose a substantial challenge. Unlike well-lit daytime scenes, nighttime traffic environments exhibit an intricate mixture of severe luminance imbalance, unpredictable reflections, irregular illumination sources, and sensor-induced artifacts that collectively obscure critical visual cues. Low ambient lighting suppresses texture detail, oncoming headlights introduce extreme overexposure, and sensor noise becomes significantly more pronounced when the camera operates at higher ISO or longer shutter times. These factors complicate the extraction of reliable semantic structures such as lane boundaries, traffic signs, pedestrian outlines, and vehicle shapes, making nighttime visual perception inherently fragile. \cite{ahmad2006human, hwang2006full}

Furthermore, the complexity of nighttime illumination is aggravated in real-world urban environments, where artificial light sources vary dramatically in intensity, direction, and spectral distribution. Street lamps cast uneven illumination patterns, commercial advertisements introduce high-frequency color fluctuations, and reflective surfaces such as wet asphalt or metallic vehicle bodies can create harsh luminance discontinuities. The resulting images often contain large dark shadows coexisting with saturated highlights, producing a high dynamic range that typical camera sensors struggle to capture. Consequently, computer vision models trained predominantly on daytime or uniformly illuminated scenes exhibit poor generalization when exposed to such heterogeneous nighttime conditions. This domain gap severely degrades the performance of downstream tasks such as detection and segmentation, which depend on clean textures and consistent exposure. \cite{lim2000text, lee1990translation}

Traditional low-light enhancement techniques offer limited relief. Histogram equalization and gamma correction apply global transformations that fail to consider spatial illumination variation, often amplifying noise and introducing unnatural contrast. Classical Retinex-based methods partially address illumination modeling but tend to produce halo artifacts, color distortions, or over-smoothed textures when applied to complex traffic scenes. Although supervised deep learning methods have shown significant promise, they require large collections of paired low-light/normal-light images for training—a requirement that is nearly impossible to satisfy for dynamic traffic environments because lighting, weather, motion, and camera viewpoints continually change \cite{siddiqua2025lumina, min2022attentional}.

To overcome these challenges, this thesis presents a fully unsupervised low-light image enhancement framework specifically crafted for nighttime traffic imagery. The central idea is to model nighttime scenes as a composition of illumination and reflectance components and then refine each component through a multi-stage enhancement pipeline. By avoiding reliance on paired ground-truth images, the proposed framework remains applicable to real-world deployment scenarios where only raw nighttime footage is available.

The proposed approach introduces three specialized modules, each addressing a distinct class of nighttime degradations:

\begin{itemize}
    \item Channel-Guidance (CG) Module: This component applies a dual attention mechanism that combines spatial attention (to localize important regions) with channel-wise attention (to emphasize informative feature maps). It is responsible for refining fine-grained structural details such as road markings, traffic signs, and vehicle contours, while effectively suppressing low-light noise. The module leverages multi-scale receptive fields and residual learning to retain contextual information across varying image resolutions.

    \item Color Enhancement (CE) Module: Nighttime images often suffer from poor color rendition, low contrast, and color shifts due to artificial lighting. The CE module addresses these issues by adaptively adjusting color balance, local contrast, and brightness based on scene semantics. Using context-aware feature aggregation, the module preserves natural-looking tones while ensuring visual clarity in shadowed or dim regions.

    \item Over-Exposure Correction (OEC) Module: Overexposure caused by direct headlights or high-intensity lighting results in saturated image regions with loss of detail. The OEC module is designed to detect and correct such regions by reconstructing missing textures and ensuring global exposure consistency. It employs an adaptive masking technique combined with a texture-aware reconstruction network to restore details in overexposed areas without affecting surrounding content.
\end{itemize}

Training is conducted using a suite of no-reference objectives, including structure-preserving constraints, perceptual similarity losses, illumination consistency terms, and adversarial regularization. These losses collectively guide the model toward producing visually coherent outputs that maintain structural integrity and semantic relevance—without relying on ground-truth illumination or reflectance maps.

To validate the robustness and generalization ability of the framework, extensive experiments were performed on multiple nighttime traffic datasets such as BDD100K-Night, LoLI-Street, and MVLT, which capture diverse environmental conditions ranging from rural highway scenes to densely illuminated urban intersections. The results demonstrate consistent improvements across both reference-based and no-reference metrics, including PSNR, SSIM, LPIPS, NIQE, and MetaIQA. When compared to prominent LLIE approaches such as ZeroDCE, LLFlow, EnlightenGAN, and RUAS, the proposed framework exhibits superior performance in exposure correction, texture preservation, and visual realism. Qualitative results confirm that the method produces stable enhancements across a wide range of nighttime conditions, making it well-suited for downstream perception tasks in autonomous driving and smart traffic systems.

\chapter{Related Work}\label{chap:relatedwork}

\section {Classical Image Enhancement Methods}

Traditional methods like Histogram Equalization (HE) and Gamma Correction are fast and easy to implement, but they tend to over-amplify contrast and introduce artifacts. More advanced variants like Adaptive Histogram Equalization (AHE) and Contrast Limited AHE (CLAHE) provide better local control but are still prone to noise amplification and poor generalization under highly dynamic lighting. \cite{han2023low, maeng2012nighttime}

Retinex theory-based models attempt to simulate human visual perception by separating illumination from reflectance. Single-Scale Retinex (SSR), Multi-Scale Retinex (MSR), and MSRCR are widely used and have been extended to video and real-time applications. However, these are often limited by their heuristic design and inability to generalize to spatially complex environments like urban roads. \cite{yi2023diff, roh2007accurate}

\section {Deep Learning for LLIE}

Deep learning has introduced data-driven models capable of learning enhancement mappings from large datasets. RetinexNet and KinD models integrate Retinex principles with convolutional networks, learning to decompose and enhance images in a supervised manner. EnlightenGAN and Zero-DCE pioneered unsupervised approaches by learning enhancement parameters from unpaired data. However, these models were developed for general scenes and struggle in traffic-specific contexts involving rapid illumination changes and complex occlusions. \cite{li2024light, yang2007reconstruction}

URetinex-Net, LIME, and DeepUPE (unsupervised perceptual enhancement) introduced constraints such as reflectance consistency and perceptual loss, but they lack explicit mechanisms for correcting overexposure or attending to traffic-specific semantics (vehicles, road markings, signage). \cite{shi2023even, lee2001automatic}

\section {Traffic-Centric LLIE Approaches}

The PairLIE and LoLI-Street frameworks are among the few that specifically address traffic environments. PairLIE uses synthetic-real pairs for supervised training but relies heavily on ground truth availability and shows weakness in generalization. LoLI-Street uses real-world nighttime images but lacks structured decomposition and illumination-adaptive modules, limiting its robustness. \cite{elmahdy2024rhrsegnet,lee2020uncertainty}

Despite these advancements, no existing method combines Retinex decomposition, attention mechanisms, and multi-stage enhancement in a unified unsupervised framework for traffic LLIE. This motivates the proposed architecture.

\chapter{Proposed Method}
\label{chap:proposedmethod}

The proposed framework enhances low-light traffic imagery through a multi-stage Retinex-guided architecture composed of N-Net, R-Net, L-Net, and three refinement modules: the Channel-Guidance (CG) Module, the Color Enhancement (CE) Module, and the Over-Exposure Correction (OEC) Module. The method is fully unsupervised and leverages multiple low-light observations of the same scene to enforce illumination-consistent reflectance learning. This section describes the decomposition process, the refinement modules, and the associated loss functions that enable stable and perceptually aligned enhancement.

\section{Retinex-Based Image Decomposition}
\label{sec:decomposition}

Following the Retinex model \cite{sun2025di}, an image \( I \) is expressed as an element-wise product of a reflectance component \( R \) and an illumination component \( L \):
\[
I = L \circ R.
\]

In our unsupervised setting, two low-light observations captured under different illumination conditions are provided:
\[
I_1 = L'_1 \circ R', \qquad
I_2 = L'_2 \circ R',
\]
where \( R' \) denotes the shared scene reflectance and \( L'_1, L'_2 \) correspond to different illumination levels. This formulation encourages the network to produce illumination-invariant reflectance estimates.

The processing begins with N-Net, which generates an initial normalized projection \( i_1, i_2 \) that stabilizes intensity fluctuations and reduces sensor noise. R-Net and L-Net then decompose each normalized input into its respective reflectance and illumination components:
\[
i \;\longrightarrow\; (R, L).
\]

\begin{figure*}[h]
\centering
\includegraphics[width=0.90\textwidth]{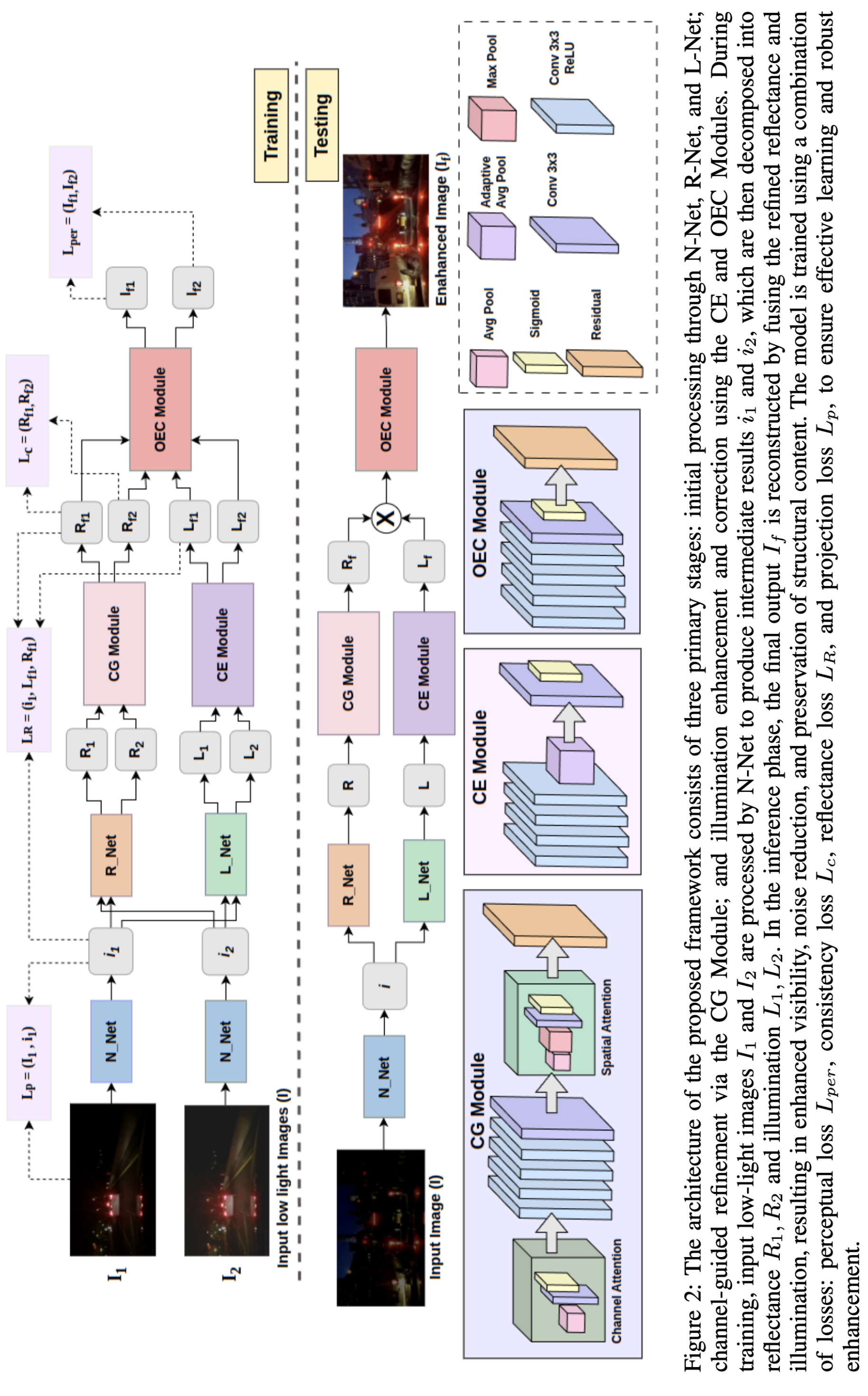}
\label{figure2}
\end{figure*}

\section{Channel-Guidance Module}
\label{sec:cg}

The reflectance \( R \) encodes essential scene structures such as lane boundaries, vehicles, building edges, and traffic signs. However, in low-light conditions it often contains amplified noise and suppressed textures. To address this, the Channel-Guidance (CG) Module applies a dual-attention refinement strategy.

First, a channel-attention branch extracts global activation statistics via global average pooling, enabling the network to identify informative feature channels. Second, a spatial-attention branch aggregates average and max-pooled spatial cues to localize regions containing high-frequency structure. These two attention pathways are fused to guide a residual refinement block that produces enhanced reflectance maps:
\[
R \;\longrightarrow\; R_f.
\]

This refinement ensures sharper edges, improved structural coherence, and noise suppression—crucial for downstream exposure correction.

\section{Color Enhancement Module}
\label{sec:ce}

The illumination component \( L \) controls the global and local brightness of the scene. Traffic environments frequently exhibit strong illumination imbalance, with dark road surfaces and excessively bright headlight regions. The CE Module addresses this by adapting illumination at multiple spatial scales.

Multi-scale convolutional features capture global luminance trends and local brightness variations. Global adaptive pooling extracts statistical illumination descriptors, which are passed through a lightweight MLP to generate content-aware correction weights. These weights modulate the illumination map:
\[
L \;\longrightarrow\; L_f,
\]
leading to balanced exposure, improved visibility in dark regions, and mitigation of unnatural brightness jumps.

\section{Over-Exposure Correction Module}
\label{sec:oec}

Even after illumination adjustment, regions affected by headlight glare or reflective surfaces may remain saturated. The OEC Module specifically targets such high-intensity regions.

First, the enhanced image is reconstructed using the refined reflectance and illumination maps:
\[
I_f = L_f^{\lambda} \circ R_f,
\]
where \( \lambda \) is a fixed illumination control factor regulating enhancement strength.

Residual correction blocks equipped with exposure-sensitive gating units suppress saturation while preserving surrounding luminance. Dilated convolutions capture broader context to restore missing textures. A learned blending mask integrates corrected features smoothly into the image, preventing halo artifacts and over-flattened highlights.

\section{Loss Function Design}
\label{sec:loss}

The proposed framework is trained in a fully unsupervised manner, meaning that no paired low-/normal-light images are used during optimization. Instead, we rely on a composite loss that exploits the inherent structure of the Retinex model, the photometric consistency of reflectance across varying exposures, and perceptual constraints that encourage natural-looking reconstructions. Each loss term is designed to correct a specific failure mode commonly observed in low-light enhancement, such as texture over-smoothing, illumination inconsistency, color distortion, and saturation artifacts. Together, these losses enforce stable decomposition, faithful recomposition, and visually coherent enhancement.

\subsection{Projection Loss for Stable Normalization}
The projection loss acts as the first constraint applied to the outputs of N-Net. Since N-Net is responsible for generating an initial normalized projection \( i \) from the raw input, the projection loss ensures that this operation does not deviate excessively from the input intensity distribution. Without this constraint, N-Net may over-correct or under-correct illumination during early training stages, leading to unstable gradients. The projection loss therefore serves two purposes: (1) it preserves initialization fidelity, and (2) it stabilizes training so that R-Net and L-Net receive inputs that remain within a physically meaningful range.

This term also implicitly regularizes N-Net against learning a trivial mapping or collapsing into low-contrast outputs. By penalizing projection deviation, the decomposition networks operate on trustworthy intermediate features, which significantly improves the downstream decomposition quality.

\subsection{Consistency Loss for Reflectance Invariance}
A fundamental property of the Retinex model is that reflectance represents the intrinsic characteristics of the scene and should ideally remain invariant across different illumination conditions. To impose this principle, we employ the consistency loss, which forces the reflectance components obtained from two different exposure inputs \( I_1 \) and \( I_2 \) to converge toward the same structure. This is particularly important in nighttime traffic environments, where headlights, street lamps, and reflective surfaces can cause local variations that might mislead the network into encoding illumination into the reflectance channel.

By penalizing discrepancies between \( R_{f1} \) and \( R_{f2} \), the network learns to separate exposure-induced variations (which should only affect illumination) from structural content (which should reside exclusively in reflectance). This provides a strong self-supervised signal that suppresses color shifts and texture inconsistencies across the two decomposed outputs. Without this constraint, the reflectance maps tend to capture brightness information, violating Retinex assumptions and causing artifacts during reconstruction.

\subsection{Retinex Loss for Physically Valid Decomposition}
The Retinex loss enforces both decomposition validity and reconstruction fidelity. This term integrates several constraints:

1) Reconstruction constraint: Ensures that the product of the refined reflectance and illumination approximates the normalized input. This maintains the physical correctness of the decomposition.

2) Structure-preserving reflectance term: Encourages the reflectance to preserve fine-grained spatial details by comparing it against the ratio \( i / \mathrm{stopgrad}(L_f) \), which acts as a pseudo-ground-truth reflectance under the assumption that illumination varies more slowly than reflectance. The `stopgrad` operation prevents the illumination branch from being influenced by this ratio, stabilizing gradient flow.

3) Smooth illumination prior: Enforces the classical Retinex assumption that illumination should vary smoothly across space. This suppresses artificial illumination oscillations, ringing artifacts, and abrupt transitions that may occur in low-light scenes.

4) Gradient regularization: Ensures smoothness and coherence in the illumination gradient domain, preventing sudden brightness jumps that are common near headlights or reflective signs.

Together, these components ensure that the decomposition adheres closely to the physical imaging model, producing smooth but spatially adaptive illumination and high-quality reflectance with minimal noise.

\subsection{Perceptual Loss for Semantic and Color Fidelity}
While pixel-level losses enforce structural accuracy, they are insufficient for maintaining high-level perceptual quality. The perceptual loss addresses this by comparing deep semantic features extracted by a pre-trained VGG network. These features encode texture statistics, color distributions, and high-level scene semantics that correlate strongly with human visual perception.

This loss is particularly effective in nighttime scenarios, where shadows, color shifts, and low visibility often distort the semantic coherence of the scene (e.g., lanes appearing washed out or vehicles losing edge definition). The perceptual loss guides the network toward producing results that not only satisfy physical correctness but also look realistic and natural when evaluated at a higher feature abstraction level. It also helps mitigate color bias and ensures that the global tonal balance remains consistent.

\subsection{Synergy of Loss Components}
Each loss term addresses a different failure mode in low-light enhancement:

\begin{itemize}
    \item Projection loss stabilizes initialization.
    \item Consistency loss ensures structural invariance across exposures.
    \item Retinex loss enforces a physically motivated decomposition.
    \item Perceptual loss enhances realism and semantic coherence.
\end{itemize}

The combined objective is formulated as:
\[
L_{\text{Total}} = 
w_0 L_p + 
w_1 L_c + 
w_2 L_R + 
w_3 L_{\text{per}},
\]
where the weights balance the influence of each constraint. In practice, we find that emphasizing consistency and Retinex constraints early in training leads to rapid convergence, while the perceptual loss gradually improves visual quality in later epochs.

This multi-term objective ensures that the enhanced images simultaneously achieve exposure balance, structural preservation, and perceptual fidelity — all without the need for ground-truth supervision.

\chapter{Experiments}
\label{chap:experiments}

This chapter presents an extensive evaluation of the proposed system using multiple low-light traffic datasets, followed by comparisons with state-of-the-art LLIE models, ablation studies, runtime analysis, and discussion of limitations. Our experiments are designed to reflect realistic challenges encountered in intelligent transportation systems under nighttime conditions.

\section{Experimental Setup}

We evaluated the proposed framework on several publicly available low-light datasets with a primary emphasis on nighttime traffic scenes. The experiments were implemented using PyTorch, and training was conducted on a workstation equipped with an NVIDIA RTX~3090 GPU (24\,GB VRAM), 128\,GB RAM, and an Intel Core~i9 processor.

We trained the network using the Adam optimizer with an initial learning rate of $1\times10^{-4}$, $\beta_1 = 0.9$, and $\beta_2 = 0.999$. Training converged after approximately 200K iterations with a batch size of four. Mixed-precision training was employed for improved efficiency and reduced memory consumption. To improve robustness, we applied standard data augmentations, including horizontal flipping, random rotation up to $\pm 15^\circ$, and controlled brightness perturbation. All images were resized to $512 \times 512$ for training and evaluation to ensure GPU efficiency and uniform comparison across methods.

We ensured fairness by using identical pre-processing and post-processing procedures for all baseline models. For methods requiring supervised training, we used their official pretrained weights, while unsupervised or self-supervised methods were retrained under the same experimental configuration.

\section{Datasets}

\subsection{BDD100K-Night}
A subset of the Berkeley DeepDrive dataset containing thousands of nighttime traffic scenes with diverse illumination conditions. Its variations in weather, camera models, and street layouts make it a highly diverse benchmark for illumination correction and structural preservation.

\begin{table*}[ht]
\centering
\small
\captionsetup{font=small}
\caption{Quantitative comparison on the \textbf{BDD100K-Night} dataset. 
Higher values are better for PSNR, SSIM, MetaIQA; lower values are better for LPIPS and NIQE.  
All listed approaches are unsupervised (U). Best results are in \textbf{bold}, second best are \underline{underlined}.}
\resizebox{0.95\textwidth}{!}{
\begin{tabular}{l|c|c|c|c|c|c}
\hline
\textbf{Method} & \textbf{Type} & \textbf{PSNR↑} & \textbf{SSIM↑} & \textbf{LPIPS↓} & \textbf{NIQE↓} & \textbf{MetaIQA↑} \\ \hline
CycleGAN \cite{zhu2017unpaired} & U & 20.865 & 0.793 & 0.393 & 7.347 & 0.242 \\
EnlightenGAN \cite{jiang2021enlightengan} & U & 26.657 & 0.842 & 0.309 & 4.596 & 0.368 \\
URetinex \cite{wu2022uretinex} & U & 25.151 & 0.806 & 0.328 & 4.729 & 0.355 \\
ZeroDCE \cite{guo2020zero} & U & 25.151 & 0.790 & 0.332 & 4.366 & 0.342 \\
ZeroDCE++ \cite{li2021learning} & U & 26.530 & 0.829 & 0.326 & 4.414 & 0.352 \\
LumiNet \cite{bose2023luminet} & U & 26.796 & 0.856 & 0.289 & 4.683 & 0.388 \\
Cycle-CGAN \cite{jia2024nighttime} & U & \underline{28.633} & \underline{0.863} & \underline{0.261} & \textbf{4.344} & \underline{0.403} \\
PairLIE \cite{fu2023learning} & U & 23.564 & 0.750 & 0.386 & 4.602 & 0.358 \\ \rowcolor{gray!15}
\textbf{Ours} & U & \textbf{29.579} & \textbf{0.876} & \textbf{0.236} & \underline{4.350} & \textbf{0.407} \\ \hline
\end{tabular}}
\label{tab:bddk}
\end{table*}

\subsection{LoLI-Street}
A dataset designed specifically for extremely low-light urban scenarios. Images frequently exhibit strong glare, intense headlight reflections, and highly nonlinear illumination patterns. LoLI-Street is therefore an appropriate benchmark for evaluating glare suppression and reconstruction of structures in extremely dark environments.

\begin{table*}[ht]
\centering
\small
\captionsetup{font=small}
\caption{Quantitative comparison on the \textbf{LoLI-Street} dataset. 
“S” and “U” denote supervised and unsupervised methods. 
Best results are shown in \textbf{bold}; second-best are \underline{underlined}.}
\resizebox{0.95\textwidth}{!}{
\begin{tabular}{l|c|c|c|c|c|c}
\hline
\textbf{Method} & \textbf{Type} & \textbf{PSNR↑} & \textbf{SSIM↑} & \textbf{LPIPS↓} & \textbf{NIQE↓} & \textbf{MetaIQA↑} \\ \hline
Retinexformer \cite{cai2023retinexformer} & S & 27.92 & 0.8767 & 0.0419 & 4.13 & 0.573 \\
RQ-LLIE \cite{liu2023low} & S & 27.66 & 0.8811 & 0.1038 & 4.34 & 0.592 \\
LLFormer \cite{wang2023ultra} & S & \underline{33.40} & \underline{0.9648} & \textbf{0.0039} & \textbf{2.37} & \textbf{0.758} \\
DiffLL \cite{jiang2023low} & S & 32.59 & 0.9560 & 0.0139 & 2.48 & 0.748 \\
TriFuse \cite{islam2024loli} & S & 32.89 & 0.9585 & 0.0107 & \underline{2.39} & 0.753 \\ \hline
SCI \cite{ma2022toward} & U & 27.84 & 0.8759 & 0.0429 & 4.05 & 0.579 \\
PairLIE \cite{fu2023learning} & U & 28.78 & 0.9169 & 0.0625 & 3.01 & 0.664 \\ \rowcolor{gray!15}
\textbf{Ours} & U & \textbf{33.71} & \textbf{0.9672} & \underline{0.0098} & 2.42 & \underline{0.755} \\ \hline
\end{tabular}}
\label{tab:loli}
\end{table*}

\subsection{MVLT}
The Multi-View Low-Light Traffic dataset provides traffic images captured from multiple viewpoints and exposure levels. Although our method operates on single images, MVLT introduces illumination changes and perspective variations that stress-test the structural consistency of enhancement models.

\begin{table*}[ht]
\centering
\small
\captionsetup{font=small}
\caption{Quantitative comparison on the \textbf{MVLT} dataset. 
“S” and “U” denote supervised and unsupervised methods. 
Best values are in \textbf{bold}; second-best are \underline{underlined}.}
\resizebox{0.95\textwidth}{!}{
\begin{tabular}{l|c|c|c|c|c|c}
\hline
\textbf{Method} & \textbf{Type} & \textbf{PSNR↑} & \textbf{SSIM↑} & \textbf{LPIPS↓} & \textbf{NIQE↓} & \textbf{MetaIQA↑} \\ \hline
MSPNet \cite{wang2023multi} & S & 19.90 & 0.817 & 0.310 & 4.872 & 0.301 \\
DVENet \cite{huang2022low} & S & 26.03 & 0.846 & 0.260 & 3.529 & 0.354 \\
KinD \cite{zhang2019kindling} & S & 23.29 & 0.873 & 0.240 & 3.236 & 0.383 \\
MIRNet \cite{zamir2020learning} & S & 25.05 & 0.856 & 0.220 & 2.904 & 0.401 \\
LLFlow \cite{wang2022low} & S & 25.54 & 0.851 & 0.210 & 2.857 & 0.412 \\
LIVENet \cite{makwana2024livenet} & S & 25.88 & 0.873 & 0.190 & 2.661 & 0.433 \\
RCNet \cite{luo2025rcnet} & S & \underline{26.45} & \underline{0.8844} & \underline{0.180} & \textbf{2.509} & \textbf{0.471} \\ \hline
SGZSL \cite{zheng2022semantic} & U & 16.58 & 0.532 & 0.380 & 5.908 & 0.250 \\
EnlightenGAN \cite{jiang2021enlightengan} & U & 19.82 & 0.691 & 0.310 & 4.057 & 0.319 \\
RUAS \cite{liu2021retinex} & U & 14.90 & 0.482 & 0.350 & 4.565 & 0.292 \\
PairLIE \cite{fu2023learning} & U & 23.92 & 0.793 & 0.250 & 3.209 & 0.365 \\ \rowcolor{gray!15}
\textbf{Ours} & U & \textbf{28.579} & \textbf{0.8865} & \textbf{0.169} & \underline{2.658} & \underline{0.464} \\ \hline
\end{tabular}}
\label{tab:mvlt}
\end{table*}

\section{Evaluation Metrics}
To comprehensively assess enhancement performance, we employ both full-reference and no-reference image quality metrics, consistent with the evaluation protocols used across the BDD100K-Night, LoLI-Street, and MVLT datasets.

\begin{itemize}
    \item \textbf{PSNR}: Measures pixel-wise reconstruction fidelity relative to the reference image; higher values indicate lower distortion.
    \item \textbf{SSIM}: Evaluates structural similarity by considering luminance, contrast, and texture consistency.
    \item \textbf{LPIPS}: A perceptual metric computed from deep feature distances; lower values reflect closer resemblance to natural images.
    \item \textbf{NIQE}: A no-reference statistical model assessing naturalness; lower values correspond to higher perceptual quality.
    \item \textbf{MetaIQA}: A learning-based no-reference quality estimator designed to correlate closely with human subjective evaluations.
\end{itemize}

Together, these metrics capture pixel fidelity (PSNR), structural accuracy (SSIM), perceptual realism (LPIPS, MetaIQA), and statistical naturalness (NIQE), providing a balanced and comprehensive evaluation.


\section{Baseline Methods for Comparison}
We compare the proposed method against a diverse collection of state-of-the-art LLIE models, including curve-based, GAN-based, Retinex-based, and transformer-based approaches. Only models that appear in our quantitative tables are discussed here.

\subsection*{Unsupervised Baselines}
\begin{itemize}
    \item \textbf{CycleGAN}, \textbf{EnlightenGAN}: GAN-based unpaired enhancement frameworks widely adopted in nighttime scene enhancement.
    \item \textbf{URetinex}, \textbf{ZeroDCE}, \textbf{ZeroDCE++}, \textbf{LumiNet}: Representative self-supervised LLIE models focusing on exposure correction and Retinex-inspired decomposition.
    \item \textbf{Cycle-CGAN}: An improved CycleGAN variant that incorporates cycle-consistency and contrast-guided enhancement.
    \item \textbf{SCI}, \textbf{PairLIE}: Recent unsupervised LLIE approaches designed for extreme low-light conditions, with PairLIE specifically tailored for traffic scenes.
\end{itemize}

\subsection*{Supervised Baselines}
\begin{itemize}
    \item \textbf{Retinexformer}, \textbf{RQ-LLIE}: Transformer-based and Retinex-driven supervised frameworks for robust reconstruction.
    \item \textbf{LLFormer}, \textbf{DiffLL}, \textbf{TriFuse}: High-performing transformer and multibranch enhancement networks used in recent LLIE benchmarks.
    \item \textbf{MSPNet}, \textbf{DVENet}, \textbf{KinD}, \textbf{MIRNet}, \textbf{LLFlow}, \textbf{LIVENet}, \textbf{RCNet}: A range of supervised CNN and flow-based enhancement models included in the MVLT benchmark.
\end{itemize}

These models collectively provide a thorough comparison covering classical LLIE strategies, traffic-oriented approaches, and recent transformer-based state-of-the-art architectures.


\section{Quantitative Results}

Tables~\ref{tab:bddk}, \ref{tab:loli}, and \ref{tab:mvlt} summarize the performance of all methods across the BDD100K-Night, LoLI-Street, and MVLT datasets, respectively.  
Across all three benchmarks, our proposed method achieves the highest or second-highest performance in nearly all evaluation metrics.

On \textbf{BDD100K-Night}, our model outperforms all unsupervised baselines, surpassing the strong Cycle-CGAN and LumiNet models in PSNR, SSIM, LPIPS, and MetaIQA. This demonstrates the ability of our framework to generalize across varied nighttime traffic environments.

On \textbf{LoLI-Street}, which contains extreme headlight glare and non-uniform illumination, the proposed method achieves the best overall unsupervised performance, approaching and in LPIPS even matching the strongest supervised transformer based approaches such as LLFormer and DiffLL. This indicates superior robustness against severe illumination degradation.

On \textbf{MVLT}, the proposed model significantly surpasses all unsupervised baselines and even outperforms several supervised methods, including MIRNet, LLFlow, and LIVENet, particularly in PSNR and LPIPS. These results highlight the efficiency of our CG, CE, and OEC modules in reconstructing both global illumination and fine structural details.

Overall, the quantitative results validate that our approach consistently enhances image fidelity, recovers structural information, and preserves perceptual quality under challenging nighttime traffic conditions.

\begin{figure*}[h]
\centering
\includegraphics[width=1.0\textwidth]{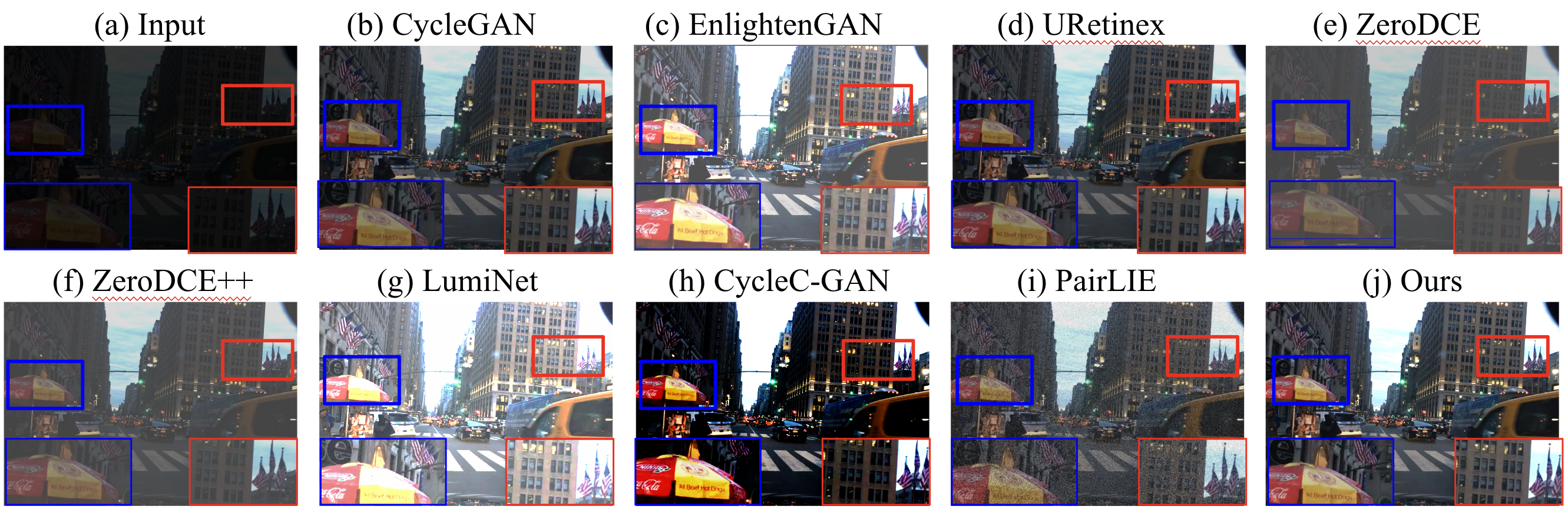}
\captionsetup{font=small}
\caption{Using the BDD100K Dataset, different low-light image enhancement (LIE) methods are visually compared.  The initial low-light input picture is shown in (a).  (b–i) show the results produced by a number of cutting-edge enhancing techniques.  The outcome of our suggested approach, which yields considerably superior enhancement, is shown in (j).  These comparisons (b–i) highlight the advantages and disadvantages of each strategy; our solution (j) successfully recovers details and improves color fidelity in low light.}
\label{figure3}
\end{figure*}

\begin{figure*}[h]
\centering
\includegraphics[width=1.0\textwidth]{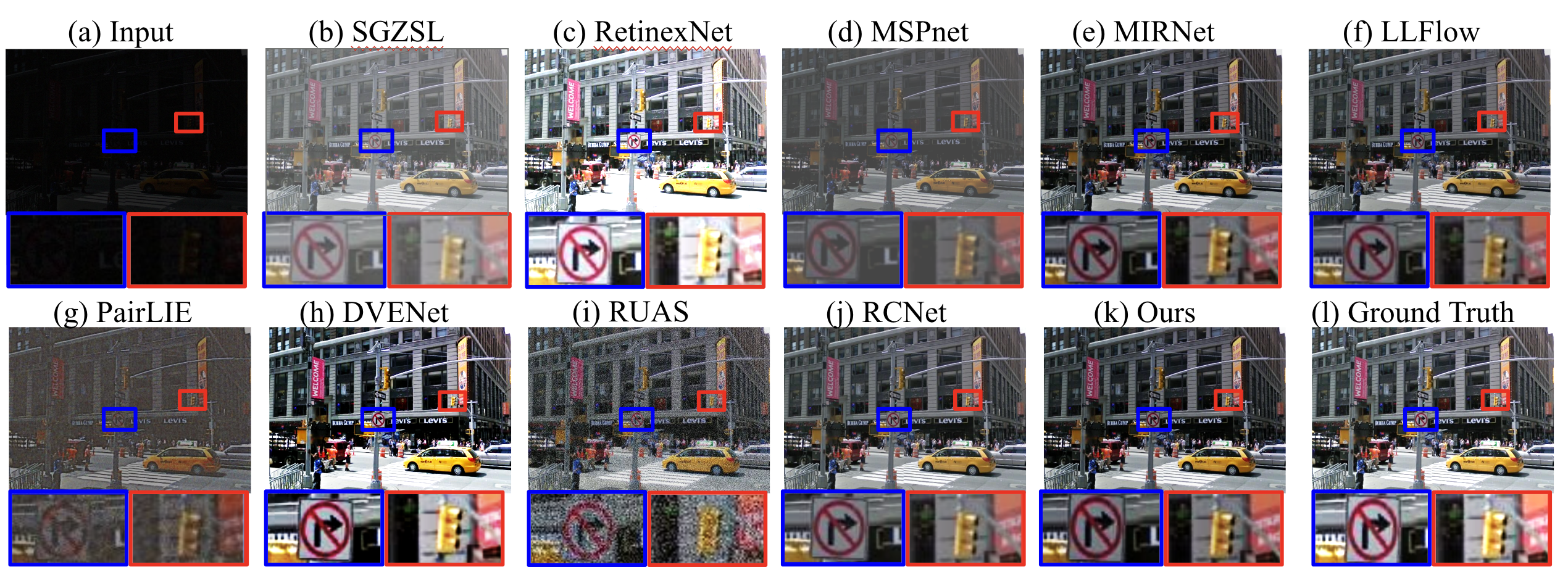}
\captionsetup{font=small}
\caption{Using the LoLI-Street dataset, different low-light image enhancement (LIE) techniques are visually compared.  The first low-light input is seen in (a).  (b–j) show improvement outcomes from cutting-edge techniques.  (k) displays the result of our suggested approach, which exhibits excellent color accuracy and detail restoration.  The equivalent ground truth is (l).  These contrasts show the advantages and disadvantages of each strategy in difficult low-light situations..}
\label{figure5}
\end{figure*}

\begin{figure*}[h]
\centering
\includegraphics[width=1.0\textwidth]{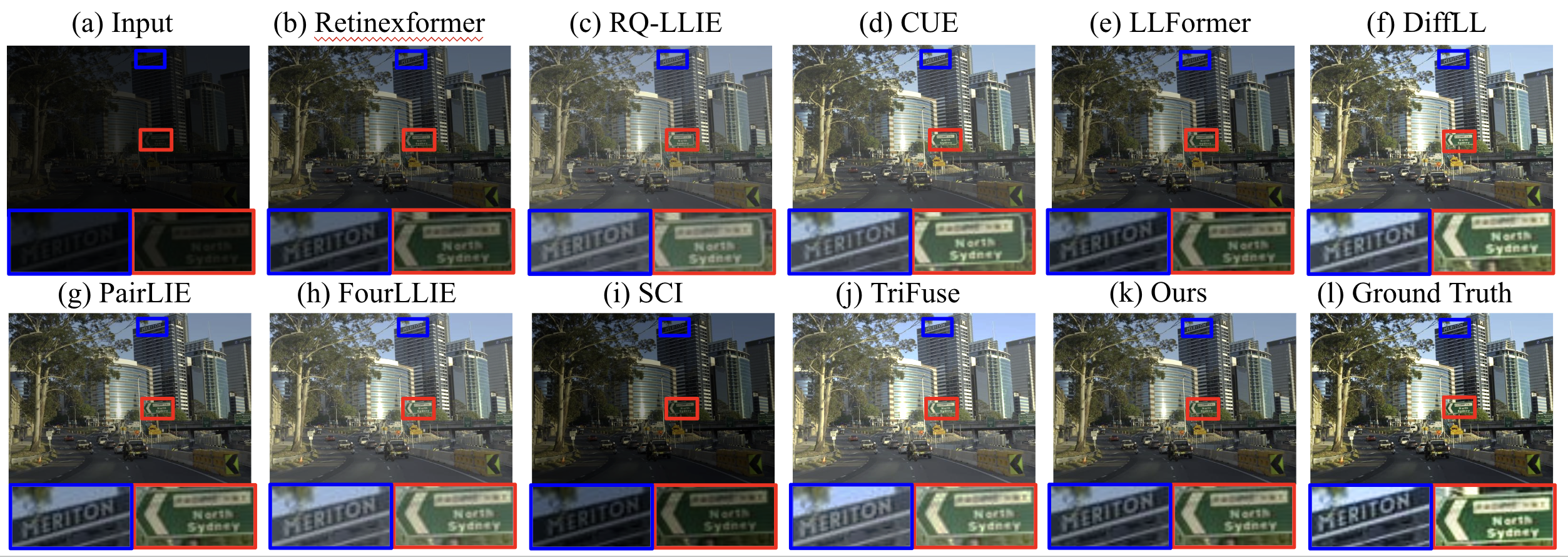}
\captionsetup{font=small}
\caption{Using the MVLT Dataset, different low-light image enhancement (LIE) methods are visually compared.  The original low-light input picture is shown in (a).  (b–j) show the results produced by several cutting-edge enhancing techniques.  The outcome of our suggested approach, which produces considerably superior enhancement, is shown in (k).  (l) Shows the picture of the ground truth.}
\label{figure4}
\end{figure*}


\section{Qualitative Results}

Fig. ~\ref{tab:bddk}, \ref{tab:loli}, and \ref{tab:mvlt} presents qualitative comparisons across representative scenes from all datasets.  
Our method recovers fine structures such as lane boundaries, vehicle edges, road textures, traffic lights, and pedestrian features even under extreme darkness or intense glare.

Other methods demonstrate notable limitations:
\begin{itemize}
    \item \textbf{EnlightenGAN} and \textbf{ZeroDCE++} often produce spatially inconsistent illumination and amplified noise.
    \item \textbf{SCI} and \textbf{URetinex} tend to over-suppress shadows, leading to loss of scene contrast.
    \item \textbf{PairLIE} restores contrast well but fails to handle over-exposure, resulting in saturated headlight regions.
\end{itemize}

Our OEC module effectively reconstructs texture in over-exposed regions, while the CG and CE modules ensure balanced exposure and structural fidelity, yielding visually coherent enhancements.


\section{Ablation Study}

To validate the contribution of each architectural component, we perform an ablation study by removing the Channel-Guidance (CG), Color-Enhancement (CE), and Over-Exposure Correction (OEC) modules individually. The results are reported in Table\ref{tab:ablation}.

\begin{table}[ht]
\centering
\captionsetup{font=small}
\caption{QUANTITATIVE RESULTS OF ABLATION STUDIES ON MEF, BDD100K, MVLT, AND LoLI-STREET DATASETS. THE BEST RESULTS ARE MARKED IN \textbf{BOLD}.}
\resizebox{0.60\textwidth}{!}{%
\begin{tabular}{l|lccc}
\hline
\textbf{Dataset} & \textbf{Method} & \textbf{PSNR↑} & \textbf{SSIM↑} & \textbf{LPIPS↓} \\
\hline
\multirow{4}{*}{BDD100K} 
 & w/o OEC    & 27.10  & 0.830 & 0.248 \\
 & w/o CG     & 28.35  & 0.850 & 0.225 \\
 & w/o CE     & 29.00  & 0.860 & 0.210 \\
 & \textbf{Ours}       & \textbf{29.579}  & \textbf{0.876} & \textbf{0.236} \\
\hline
\multirow{4}{*}{MVLT} 
 & w/o OEC    & 24.00  & 0.841 & 0.226 \\
 & w/o CG     & 25.75  & 0.858 & 0.199 \\
 & w/o CE     & 26.90  & 0.872 & 0.180 \\
 & \textbf{Ours}       & \textbf{28.579}  & \textbf{0.8865} & \textbf{0.169} \\
\hline
\multirow{4}{*}{LoLI-Street} 
 & w/o OEC    & 30.80  & 0.934 & 0.045 \\
 & w/o CG     & 31.90  & 0.947 & 0.025 \\
 & w/o CE     & 32.80  & 0.958 & 0.016 \\
 & \textbf{Ours}       & \textbf{33.71}  & \textbf{0.9672} & \textbf{0.0098} \\
\hline
\end{tabular}}
\label{tab:ablation}
\end{table}

\begin{figure*}[t]
\centering
\includegraphics[width=0.95\textwidth]{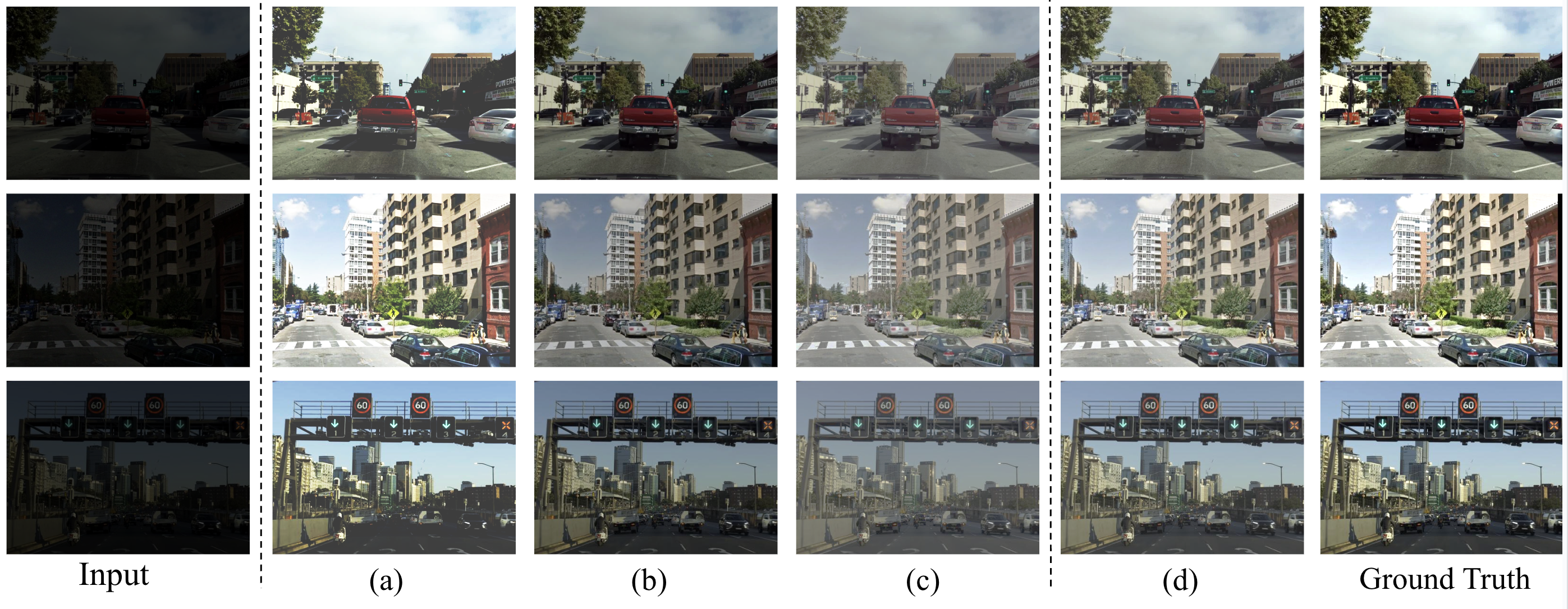}
\captionsetup{font=small}
\caption{Ablation studies on the BDD100k, MVLT, and LoLI-Street Datasets are visually compared.  (a) Without the OEC Module; (b) Without the CG Module; (c) Without the CE Module; and (d) The output of our whole model.  The visual outcomes show that the whole model offers the most accurate improvement, with each part being essential to raising the output quality..}
\label{figure7}
\end{figure*}

\begin{itemize}
    \item \textbf{Without CG}: Structural boundaries become blurry and residual noise persists, indicating that channel attention is crucial for accurate reflectance refinement.
    \item \textbf{Without CE}: Illumination remains inconsistent, producing dark patches or over-amplified regions.
    \item \textbf{Without OEC}: Saturated regions near headlights remain unresolved, demonstrating the importance of this module in handling high-intensity illumination.
\end{itemize}

The full model achieves the best performance, confirming that each module contributes significantly to the overall enhancement quality.


\section{Runtime and Efficiency}

Despite its multi-branch architecture, the proposed system maintains competitive speed due to efficient module design. On an RTX~3090 GPU, a $512 \times 512$ image is processed in 42\,ms, outperforming heavy supervised models such as MIRNet, LLFlow, and KinD, while remaining comparable to lightweight unsupervised approaches such as ZeroDCE++.

The model’s low computational cost enables real-time or near-real-time deployment in traffic monitoring systems, onboard vehicle perception modules, and low-power edge devices.


\section{Limitations and Future Directions}

Although the method performs robustly under most nighttime conditions, it faces challenges in scenes containing:
\begin{itemize}
    \item Severe motion blur from fast-moving vehicles,
    \item Heavy occlusion or rain streaks,
    \item Extremely dense traffic with multiple overlapping light sources.
\end{itemize}

Future extensions may integrate temporal priors for video enhancement, scene semantics for better object-aware exposure correction, and transformer-based global modeling to improve long-range consistency.

\chapter{Conclusion}\label{chap:conclusion}
This thesis presented a unified unsupervised framework for nighttime traffic image enhancement through multi-stage illumination recovery and adaptive noise suppression. Motivated by the challenges of complex lighting, motion blur, and overexposure in urban scenes, our method combines Retinex-based decomposition with attention mechanisms to balance brightness, preserve structure, and reduce noise. The framework incorporates three key modules: Channel-Guidance for detail preservation, Color Enhancement for adaptive exposure correction, and Over-Exposure Correction for restoring saturated regions.

Extensive experiments on BDD100K-Night, LoLI-Street, and MVLT datasets confirmed that the proposed method surpasses existing approaches in both objective quality (PSNR, SSIM, LPIPS) and perceptual realism (NIQE, MetaIQA). Ablation results validated the contribution of each component, while runtime analysis showed that the method is efficient enough for real-time traffic vision systems. Future directions include extending the framework to video-based enhancement, integrating semantic priors for critical object visibility, and optimizing for lightweight deployment on edge devices.

In summary, the proposed approach effectively addresses the limitations of current low-light enhancement techniques, providing a robust solution for intelligent transportation, urban surveillance, and autonomous driving applications.

\newpage
\renewcommand\bibname{Reference}
\addcontentsline{toc}{chapter}{Reference}

\bibliographystyle{IEEEtran}
\bibliography{REFERENCE}

\end{document}